\documentclass[11pt,a4paper]{article}
\usepackage[latin9]{inputenc}
\usepackage{amsmath}
\usepackage{graphicx}
\PassOptionsToPackage{normalem}{ulem}
\usepackage{ulem}

\makeatletter

\pdfpageheight\paperheight
\pdfpagewidth\paperwidth

\@ifundefined{date}{}{\date{}}
\usepackage[hyperref]{acl2019}
\usepackage{times}
\usepackage{latexsym}

\usepackage{url}
\usepackage{algorithm}
\usepackage{algpseudocode}
\usepackage{multirow}
\usepackage{amsmath,amssymb}

\aclfinalcopy 

\title{Massive Styles Transfer with Limited Labeled Data}

\author{Hongyu Zang \and Xiaojun Wan\\
Institute of Computer Science and Technology, Peking University \\
The MOE Key Laboratory of Computational Linguistics, Peking University \\ \texttt{\{zanghy, wanxiaojun\}@pku.edu.cn}}

\makeatother

\begin{document}
\maketitle 
\begin{abstract}
Language style transfer has attracted more and more attention in the past few years. Recent researches focus on improving neural models targeting at transferring from one style to the other with labeled data. However, transferring across multiple styles is often very useful in real-life applications. Previous researches of language style transfer have two main deficiencies: dependency on massive labeled data and neglect of mutual influence among different style transfer tasks. In this paper, we propose a multi-agent style transfer system (MAST) for addressing  multiple style transfer tasks with limited labeled data, by leveraging abundant unlabeled data and the mutual benefit among the multiple styles. A style transfer agent in our system not only learns from unlabeled data by using techniques like denoising auto-encoder and back-translation, but also learns to cooperate with other style transfer agents in a self-organization manner. We conduct our experiments by simulating a set of real-world style transfer tasks with multiple versions of the Bible. Our model significantly outperforms the other competitive methods. Extensive results and analysis further verify the efficacy of our proposed system.
\end{abstract}

\section{Introduction}

There are various language styles in our life. For example, people with different age and background or in different areas talk in different ways; famous writers have their own special writing styles; network language is more fashionable than formal language, and so on. The techniques of style transfer can be applied in real life to help with generating robotic instructions \cite{DBLP:conf/emnlp/KiddonPZC15}, simplification for children, and expression personalization \cite{lin-walker:2017:StyVa}. It is often very useful to transfer from one given style to a number of different styles, thus promoting text generation applications to meet specific requirements for different target audiences.

Most existing style transfer methods \cite{jhamtani-EtAl:2017:StyVa,ficler-goldberg:2017:StyVa} heavily rely on large-scale labeled (or paired) data for training. However, the large labeled dataset across different styles are usually not easy to get. Although the data shortage problem has been alleviated by relevant researches \cite{xu:2017:StyVa}, the proposed datasets are still limited in both scales and domains. Note that unlabeled data are abundant and easy to collect for any style, which can be leveraged to enhance the style transfer models. 

Besides, different styles have internal relationships. Like Shakespeare's plays and modern literature, they differ a lot in morphology, grammar, and so on, but they still share many common expressions. Therefore, we argue that the style transfer tasks for different style pairs have mutual influences on each other. Nevertheless, the style transfer tasks for different style pairs are investigated independently in previous works, ignoring the mutual influences among them.

In this work, we investigate the problem of transferring across multiple language styles, as described below: 

\begin{quote}
    \small
    \textit{There is a set of $n$ writing styles $S=\{s_i | i$$\in [1,n]\}$ in the same language (e.g., English). For each style $s_i$, there are plenty of unlabeled data $X_i^U$. While there are also a few labeled data $(X_i^{L_{i,j}}, X_j^{L_{i,j}})$ between any two styles $s_i$ and $s_j$. The scale of labeled data is very limited due to annotator resources and time. The goal is to find a set of style transfer models $F=\{f_{i,j} | i,j\in[1,n]\}$ to transfer text data across styles.}
\end{quote}

First, we leverage unlabeled data to improve style transfer across each style pair (i.e., one-to-one style transfer) by proposing a semi-supervised framework to enhance both encoders and decoders, which is inspired by recent researches in NMT \cite{DBLP:journals/corr/abs-1711-00043,DBLP:journals/corr/abs-1710-11041}.

Second and more importantly, for multiple (or one-to-many in our setting) style transfer, the data are further used for improving those one-to-one models in a multi-agent manner and making them achieve better performances. Although this can be done in a popular multi-task learning manner \cite{DBLP:conf/icml/CollobertW08} by sharing parameters for different one-to-one transfer models. Our experiments demonstrate that parameter sharing will easily lead to performance dropping due to inconsistency of multiple tasks. Instead, we propose a multi-agent system to address the multiple style transfer problem. The one-to-one style transfer models are regarded as our basic style transfer agents. And then we design self-organization algorithms to let the agents find and use helpful neighbors to improve themselves.

We try to set up a general circumstance following the definition of the problem we come up with. We use the dataset consisting of several versions of the Bible \cite{DBLP:journals/corr/abs-1711-04731}. Without loss of generality, we set one version as the source version, while the others are target versions. As different Bible versions are translated by different authors and targeting at different crowds, the writing styles of them are different from each other. Agents for different one-to-one style transfer tasks are trained independently and then enhanced by a few neighbor agents. The evaluation results show the efficacy and superiority of our system for multiple style transfer.

The contributions are summarized as follows:

\begin{itemize}
    
    \item To the best of our knowledge, we are the first to introduce multi-agent learning to style transfer tasks.

    \item Our proposed system can leverage unlabeled data and mutual benefits across different style transfer tasks to address the data shortage problem. And our system performs significantly better than the state-of-the-art models proposed by previous works. 
    
    \item Our codes and dataset will be released at GitHub\footnote{https://github.com/zhyack/MAST}. Experiments can be easily reproduced and extended.

\end{itemize}

\section{Related Works}

Among researches on all kinds of style transfer tasks, our work is most relevant to the studies of writing style transfer \cite{jhamtani-EtAl:2017:StyVa,ficler-goldberg:2017:StyVa,DBLP:conf/acl/NisioiSPD17,DBLP:conf/acl/TsvetkovBSP18}. Xu \shortcite{xu2012paraphrasing} proposes a dataset of style transfer between Shakespeare's scripts and modern English. They build language models and train Moses decoder \footnote{http://www.statmt.org/moses} to learn paraphrasing for styles. Jhamtani \shortcite{jhamtani-EtAl:2017:StyVa} follows this work and proposes sequence-to-sequence models trained with the parallel data in the dataset. Recently, Carlson \shortcite{DBLP:journals/corr/abs-1711-04731} proposes a more content-rich dataset for similar studies, the Bibles. Sequence-to-sequence models are applied on different versions of the Bibles and get better results than Moses.

The need to leverage unlabeled data draws a lot of interests of NMT researchers. Researches like \cite{zhenyang2018,DBLP:journals/corr/abs-1711-00043}, \cite{DBLP:conf/acl/SennrichHB16}, and \cite{DBLP:journals/corr/abs-1710-11041} propose methods to build semi-supervised or unsupervised models. However, these techniques mainly designed for NMT tasks, and they haven't been widely used for style transfer tasks. Some unsupervised approaches \cite{DBLP:conf/nips/ShenLBJ17,DBLP:conf/nips/YangHDXB18} try addressing style transfer problems by using GAN \cite{goodfellow2014generative}. But their architecture shows drawbacks in content preservation \cite{DBLP:journals/corr/abs-1805-05181}. In this paper, we follow the ideas of Sennrich's \shortcite{DBLP:conf/acl/SennrichHB16} work to propose a semi-supervised method for leveraging unlabeled data of both source side and target side.

The core inspiration for our proposed system comes from the idea of multi-agent system design. A P2P self-organization system \cite{Gorodetskii2012} have been successfully applied in practical security systems. They design policies for agents to choose useful neighbors to produce better predictions. It enlightens us to build style transfer systems. Researches on reinforcement learning in text generation tasks \cite{DBLP:conf/aaai/YuZWY17} also show the practicability to regard text generation models as agents with a large action space.

\section{One-to-One Style Transfer}
We use the popular attentional sequence-to-sequence models as baselines and build our system based on them.
\subsection{Attentional Sequence-to-Sequence Model (AttS2S)}

Some techniques have been applied to the vanilla sequence-to-sequence model \cite{sutskever2014sequence} to get better performance, such as bidirectional encoder \cite{schuster1997bidirectional} and attentional decoder \cite{bahdanau2014neural,DBLP:conf/emnlp/LuongPM15}, using LSTM  \cite{hochreiter1997long} instead of original RNN cells. In this work, we use the model with bidirectional encoder and Luong attentional decoder \cite{DBLP:conf/emnlp/LuongPM15}, one of the state-of-the-art models of neural language style transfer used in previous works \cite{xu2012paraphrasing,jhamtani-EtAl:2017:StyVa,DBLP:journals/corr/abs-1711-04731}, as one of our baseline models.

\subsection{Semi-Supervising with Unlabeled Data (Semi)}
Many studies in the NMT area try to make use of unlabeled data. Semi-supervised \cite{DBLP:journals/corr/abs-1710-11041} and unsupervised \cite{DBLP:journals/corr/abs-1711-00043,zhenyang2018} models are proposed to enhance the basic sequence-to-sequence models, which are very inspiring for improving the performance of style transfer models.

Among these studies, Sennrich \shortcite{DBLP:conf/acl/SennrichHB16} proposes two effective methods. One is to provide monolingual data with dummy source sentence, which is very similar to the technique of denoising auto-encoder(DAE)\cite{DBLP:conf/nips/BengioYAV13}. And the other one is to provide synthetic source data obtained from a target-to-source translation model (back-translation) to pair with the unlabeled target data. We improve this model to fit style transfer environment, making use of the unlabeled data from both source side and target side.

The semi-supervised sequence-to-sequence model (Semi) is shown in Figure 1. $Enc$ and $Dec$ are short for encoder and decoder, and $\sim$ means the data is noised. Our aim is to get $Enc_i$ and $Dec_j$, so that we can encode the source language from style $i$, and decode the embedding to the target language of style $j$. Different from the model in \cite{DBLP:conf/acl/SennrichHB16}, we use back-translation to translate the target-side outputs to source-side, which can make use of the unlabeled source data in an auto-encoder manner. In a word, there are three routes for training in the Semi model:

\begin{figure}[t!]
    \centering
    \includegraphics[width=3in]{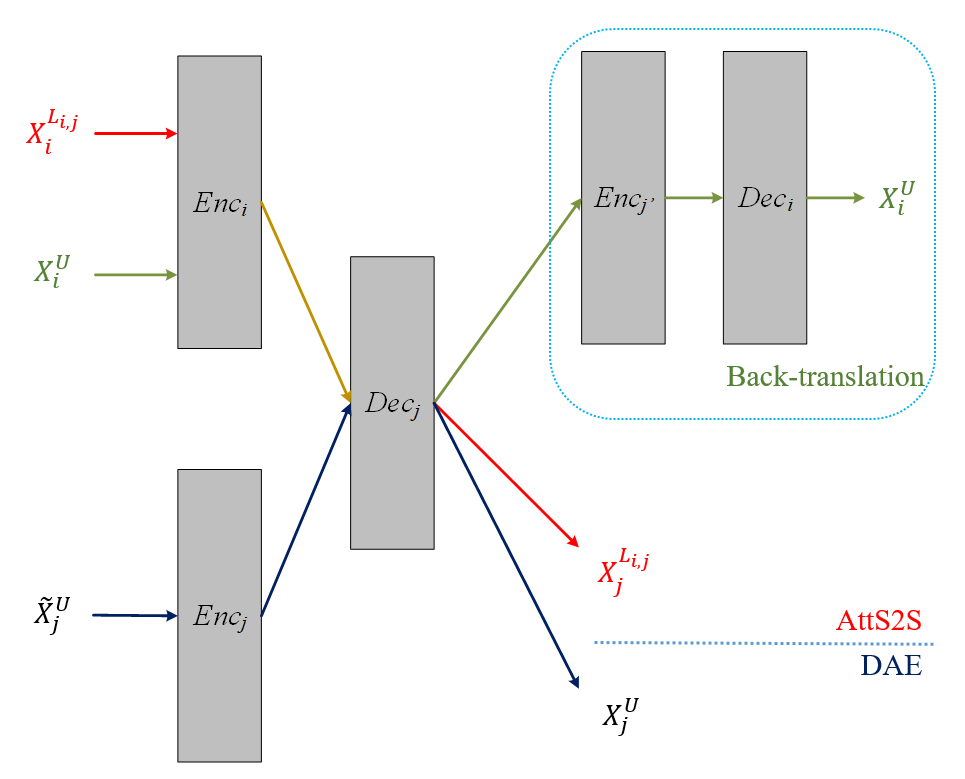}
    \caption{The structure of Semi model transferring style $i$ to style $j$. It integrates back translation and DAE with AttS2S to get better $Enc_i$ and $Dec_j$.}
    \label{figures11}
\end{figure}

\begin{itemize}
    \item{AttS2S: $X_i^{L_{i,j}}\rightarrow Enc_i \rightarrow Dec_j $$\rightarrow X_j^{L_{i,j}}$}
    \item{Back-translation: $X_i^U\rightarrow Enc_i \rightarrow Dec_j \rightarrow Enc_{j'} \rightarrow Dec_i \rightarrow X_i^U$}
    \item{DAE: $\tilde{X}_j^U\rightarrow Enc_j \rightarrow Dec_j \rightarrow X_j^U$}
\end{itemize}

We jointly train the model by randomly choosing one from the three routes for each training batch. And finally, $Enc_i$ and $Dec_j$ are enhanced by training with unlabeled data through DAE and back-translation.

\section{Multiple Style Transfer}

As we mentioned before, intuitively applying multi-task learning methods will hurt the performance on some tasks due to inconsistency in various different targets. Therefore, we propose a new approach to integrate models for different tasks not by sharing parameters, but by designing a multi-agent system. We regard the multiple Semi models as the basic agents for style transfer between the fixed source style and multiple target styles. To model how they communicate and use the information of the others, we follow the framework of the classic multi-agent system to build our multi-agent system for style transfer (MAST). There are two core steps for each agent in MAST: finding other helpful agents as neighbors and learning to predict with the neighbors.

\subsection{Self-Organization by Similarities (SOS)}
In MAST, an agent cannot make use of the information of all other agents, for some agents may be not helpful to this specific one and the computing resources are limited. Therefore, we need to design algorithms for an agent to automatically locate the most helpful agents as its neighbors. To achieve this, we propose and combine two novel strategies.

\begin{figure*}[ht!]
    \centering
    \includegraphics[width=6in]{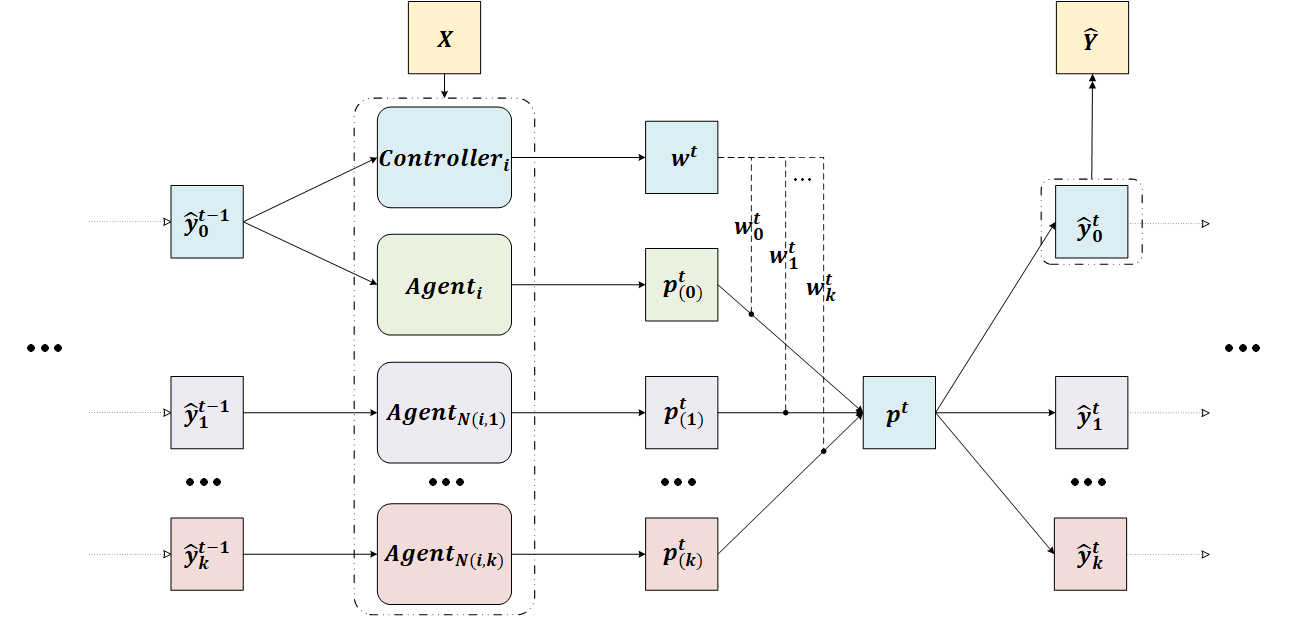}
    \caption{The training framework of MAT. The process presented here is the enhancing procedure of $Agent_i$ (the agent converts the source style to the $i$-th target style) at time step $t$.}
    \label{figures2}
    \end{figure*}

As similar styles usually share many common things, agents building on similar styles can be referential for each other. A binary classification model is an effective way to evaluate the similarity between two styles. If two styles are very different, a good classification model can easily distinguish them and produce high-accuracy predictions, while the accuracy may be much worse when the styles are very similar to each other. Therefore, for each pair of target styles, we use a binary classification model with an attentional RNN to get a representation vector. And then we use a fully connected layer for binary classification. The attention mechanism is used to distribute weights to the vectors outputted by the RNN to construct a better feature vector for binary classification. We define $ACC \in {\rm I\!R}^{n \times n}$ to include the classification results of all the target style pairs, where $n$ is the total number of target styles in the system. Then we re-scale \footnote{https://en.wikipedia.org/wiki/Feature\_scaling} $ACC$ to $ACC'$ with respect to each row (style) to make sure all the scores are within [0,1]. And we define the similarity of styles $i$ and $j$ as $SIMI_{i,j} = 1-ACC'_{i,j}$.

Apart from the similarities drawn from the classification, some agents may not perform well on their own target styles, and these agents may not provide help to a specific agent. Figuratively speaking, working with someone really unskilled or sharing no interest with you is usually a bad choice. Inspired by this, we realize that the similarity between agents and the agent performance should be balanced to make the final decision. For evaluating the performance of the agents, we use the BLEU scores \cite{papineni2002bleu} achieved on the development set. We also re-scale the BLEU scores for all agents to $[0,1]$ to get the performance scores $PERF \in {\rm I\!R}^{n}$ .

Finally, we linearly combine the similarity scores and the performance scores by Eq.1, where $\alpha$ is a coefficient weight and $SC \in {\rm I\!R}^{n \times n}$. In this work, we set $\alpha$ to 0.5 to combine the scores equally. The $k$ neighbors of agent $i$ are chosen by checking the top-$k$ scores (except for $SC_{i,i}$) in $SC_i$. It is notable that $SC_{i,j} \ne SC_{j,i}$, thus the choosing of neighbors may be unidirectional.
\begin{equation}
SC_{i,j} = \alpha*SIMI_{i,j}+(1-\alpha)*PERF_j
\end{equation}

\subsection{Multi-Agent Training (MAT)}
After setting the neighbors, we need to give an agent the ability of learning to make use of their neighbors. Imagine that you encounter some life choices. The first thing you will do is probably to ask for advice from families and friends. And then you will weigh all the suggestions according to your estimate of the situation to make your final choice. In a similar way, we propose the multi-agent training framework - MAT as illustrated in Figure 2.

In Figure 2, $Agent_i$ is the $i$-th agent in the system we want to enhance. We denote $N(i,j) (j \in [1,k])$ as the indexes of its $k$ neighbors. All these basic agents are pre-trained by using the Semi model. Apart from the basic agents, an auxiliary model $Controller_i$ for $Agent_i$ is trained, which is the key to let $Agent_i$ make use of its neighbors. As each of the agents will produce a probability distribution $p_{(j)}^t (j \in [0,k])$ over its action space (vocabulary) $A_j (j \in [0,k])$ at time step $t$, $Controller_i$ is trained to predict a weight distribution $w^t$ on all the basic agents. That is to say, $Controller_i$ does not learn to make generative actions, but learns to integrate the probability distributions of all the basic agents as Eq.2 according to the environments (inputs $X$ and the last predicted word $\hat{y}_0^{t-1}$). In Eq.2, $p^t$ is over a global action space, which is the union set of $A_j (j \in [0,k])$. $M$ is a mapping operation to set the probabilities of words not in $A_j$ to $0$ when forwarding $p_{(j)}^t$ to $p^t$. Finally, predictions $\hat{y}_j^t (j \in [0,k])$ are greedily sampled from $p^t$ to let the agents keep pace with each other and get similar states at next time steps.

\begin{equation}
p^t = \displaystyle\sum_{j=0}^{k}w_{j}^t*M(p_{(j)}^t)
\end{equation}

It is noteworthy that different agents have different action spaces (vocabularies) $A_j (j \in [0,k])$. Therefore, we use $M$ to map the vector from local action spaces $A_j (j \in [0,k])$ to global action space. And the predictions  $\hat{y}_{(j)}^t (j \in [0,k])$ can be different for different agents, because words not in $A_j$ are not valid actions of $Agent_{N(i,j)}$.

In $Controller_i$, we use an RNN encoder to model the environment and another RNN decoder to predict weights. As all the basic agents are pre-trained, we fix the parameters of them. That is to say, in Figure 2, only $Controller_i$ is trained. Algorithm 1 shows the whole training process.

In MAST, training SOS and MAT is efficient. And compared with other approaches, all the modules in our system (training basic agents, SOS and MAT) can be easily deployed in a distributed manner. That is to say, when new styles get involved, it is easy and efficient to extend the system.

\begin{algorithm}[ht!]
\caption{MAT for the whole system}
\begin{algorithmic}[1]
    \Require \textit{Labeled Training Data}; \textit{Pre-trained Models}; \textit{Action Spaces (Dictionaries)}: $A_{0..k}$; \textit{Neighbor indexes}: $N \in {\rm I\!R}^{n \times k}$. \Comment{$n$ is the number of target styles, $k$ is the number of neighbors for each agent.}
    \Ensure \textit{Controllers}: $Controller_{1..n}$.

    \For{$i \leq n$}
    \State Get pre-trained models $Agent_i$ , $Agent_{N(i,j)}$  \Comment{$j \in [1,k]$}
    \State Randomly initialize $Controller_i$
    \For{each data pair $(X,Y)$ for style transfer $s_{source} \rightarrow s_i$}
    \State $Loss \gets 0$
    \State $\hat{y}_j^0 \gets \textit{\textbf{BOS} \scriptsize{(Begin Of Sentence)}}$ \Comment{$j \in [0,k]$}
    \For{$t \leq |Y|$}
    \State $p_{(0)}^t \gets Agent_i(X,\hat{y}_0^{t-1})$
    \State $p_{(j)}^t \gets Agent_{N(i,j)}(X,\hat{y}_j^{t-1})$ \Comment{$j \in [1,k]$}
    \State $w^t \gets Controller_i(X,\hat{y}_0^{t-1})$
    \State $p^t \gets \displaystyle\sum_{j=0}^{k}M(w_{j}^t*p_{(j)}^t)$
    \State $Loss \gets Loss-log(p^t_{y_0^t})$
    \State $\hat{y}_j^t \gets argmax_a(p^t_a)$ \Comment{$a \in A_j$ and $j \in [0,k]$}
    \EndFor
    \State Update $Controller_i$ to minimize $\frac{Loss}{|Y|}$
    \EndFor
    \EndFor

\end{algorithmic}
\end{algorithm}

\section{Experiments}

\subsection{Dataset}

Bible is one of the books that have the most translations. The number of English versions is over 50.  BibleGateway collects these bibles texts\footnote{https://www.biblegateway.com/. Information and differences of different bible versions can be found on this site.}. Carlson \shortcite{DBLP:journals/corr/abs-1711-04731} collates data crawled from this site and makes alignment between sentences across different versions according to chapters and sentence orders. They release 6 versions (ASV, BBE, DARBY, DRA, WEB, YLT) that are available in public domain and their preprocessing system\footnote{https://github.com/keithecarlson/Zero-Shot-Style-Transfer}. We crawl another 10 versions (AMP, CJB, CSB, ERV, ESV, KJ21, MEV, NCV, NIV, NOG) from BibleGateway and pre-process the data in the same way.


Without loss of generality, we set ASV as the source version, and the other 15 versions as target versions. For each source-target version pair, we can averagely get 29803 aligned sentence pairs. To simulate the situation with few labeled data and abundant unlabeled ones, we sample 2000 aligned pairs as the training set, 500 aligned pairs as the development set, 500 aligned pairs as the test set, and 20000 sentences in each version as unlabeled supplement materials.

\subsection{Training Details}
All the models except SOS adopt 2-layer bidirectional LSTM encoder and 2-layer attentional LSTM decoder. In SOS, only 1-layer LSTM is used. Dropout rates are set to 0.3 for all RNN layers. The dimension sizes of hidden vectors and embedding vectors are all set to 500. Parameters are optimized by stochastic gradient descent with a learning rate of 1.0.

AttS2S models are trained on the labeled training set, while Semi models also leverage the supplement unlabeled data besides the labeled ones. And for MAST, we take Semi models as pre-trained agents and use the labeled data to train the controllers. For SOS, the training set is used to train the classification model, and the development set is used to calculate similarities and performances.

\subsection{Model Comparison}
Here we present the results on five style transfer tasks (public bible versions). Baseline models are \textbf{AttS2S} and \textbf{Semi}. The results of the state-of-the-art unsupervised approach \textbf{STTBT}\footnote{https://github.com/shrimai/Style-Transfer-Through-Back-Translation}\cite{DBLP:conf/acl/TsvetkovBSP18} is also provided to compare. We also realize a multi-task model \textbf{MulT}, where models of different tasks share the parameters except for output layers. Two different ways of choosing neighbors, randomly choosing and SOS, are applied to MAST as \textbf{MAST:Rand-k} and \textbf{MAST:SOS-k}, where $k$ stands for the number of neighbors. In this comparison, we set $k$ to 2.

We use BLEU \cite{papineni2002bleu} for automatic evaluation. BLEU is widely used for evaluating the quality of the output texts according to given references. Usually, the higher BLEU scores mean better results. The results of automatic evaluation are displayed in Table 1.

As we can see in Table 1, fully unsupervised method STTBT performs really poor compared to others, although there are only a few labeled data pairs used in other methods. The results of STTBT and AttS2S can be regarded as the bottomlines of semi-supervised methods. MulT seems to achieve some improvements at most of the tasks compared to AttS2S. But we observe a serious performance dropping on ASV-WEB task. We think it is caused by the inconsistency between WEB and other versions. Sharing parameters in a multi-task learning manner amplifies the inconsistency and causes worse results than AttS2S. And the results of MulT cannot even beat Semi, which is more relevant and more comparative to the other methods. Therefore, we will not deploy any further experiment on STTBT and MulT.

We further conduct human evaluation on Amazon Mechanical Turk (AMT\footnote{https://www.mturk.com}) to make more convincing comparisons of four typical models. For each style (version) transfer task, we randomly choose 30 samples from test set and present the input texts, the generated output texts of different models and the gold output texts to three judges. Each judge is required to rate the generated texts of all models for each sample in three aspects: \textbf{fluency}, \textbf{accuracy} and \textbf{style} in a 5-pt Likert scale\footnote{https://en.wikipedia.org/wiki/Likert\_scale}. \textbf{Fluency} indicates whether the output text is fluent and grammatical. \textbf{Accuracy} indicates whether the output text is semantically consistent with the input text. And \textbf{style} indicates how the output text is likely from the target version. We require the judges to have read at least one version of the Bibles, so that they can understand the contents and find differences between versions. The average scores and statistical significance analysis are presented in Table 2. And we also present some samples from the test set in Figure 3.

\begin{table}[t!]
\begin{center}
\resizebox{3in}{0.5in}{
\begin{tabular}{|c|c|c|c|c|c|c|}
\hline\multirow{2}*{\bf  Version }&\multirow{2}*{\bf  STTBT }&\multirow{2}*{\bf  AttS2S }&\multirow{2}*{\bf  MulT }&\multirow{2}*{\bf   Semi }&\bf  MAST:&\bf  MAST:\\~&~&~&~&~&\bf  Rand-2&\bf   SOS-2\\\hline
\bf  BBE &  7.20 & 15.57 & 19.57 & 19.70 & 24.13 & \bf 24.72\\\hline
\bf  DARBY &  10.58 & 29.19 & 29.92 & 32.60 & 35.34 & \bf 37.36\\\hline
\bf  DRA &  7.40 & 15.99 & 21.22 & 21.25 & \bf 26.14 & 23.04\\\hline
\bf  WEB &  10.73 & 40.50 & 36.82 & \bf 41.94 & 38.20 & 41.08\\\hline
\bf  YLT &  7.02 & 15.08 & 17.26 & 19.03 & 20.16 & \bf 22.99\\\hline
\bf  Average &  8.59 & 23.27 & 24.96 & 26.90 & 28.79 & \bf 29.84\\\hline
\end{tabular}
}
\end{center}
\caption{\label{tables4} BLEU evaluation results of six models in five style (version) transfer tasks given ASV as the source version (the last row is the average results).}
\end{table}

\begin{table}[t!]
\begin{center}
\small
\begin{tabular}{|c|c|c|c|}
\hline
\bf \small  Model &\bf \small Fluency &\bf \small  Accuracy &\bf \small  Style \\\hline
\bf \small AttS2S & 2.65 $\,\,\,\,\,\,$ & 2.53 $\,\,\,\,\,\,\,\,\,\,$ & 2.56 $\,\,\,\,\,$ \\\hline
\bf \small Semi & 2.94 $\dagger \,\,\,$ & 2.84 $\dagger \,\,\,\,\,\,\,$ & 2.82 $\dagger \,\,$ \\\hline
\bf \small MAST:Rand-2 & 3.18 $\dagger \ddagger$ & 3.06 $\dagger \ddagger$ $\,\,\,$ & 3.02 $\dagger \ddagger$ \\\hline
\bf \small MAST:SOS-2 & \textbf{3.26} $\dagger \ddagger$ & \textbf{3.20} $\dagger \ddagger \wr$& \textbf{3.15} $\dagger \ddagger$ \\\hline

\end{tabular}
\end{center}
\caption{\label{tables5} Human evaluation results. $\dagger$, $\ddagger$, and $\wr$ are used to mark the results significantly better ($p<0.01$ in a two-tail t-test.) than \textbf{AttS2S}, \textbf{Semi}, and \textbf{MAST:Rand-2} respectively.}
\end{table}

\begin{figure}[t!]
    \centering
    \includegraphics[width=3in]{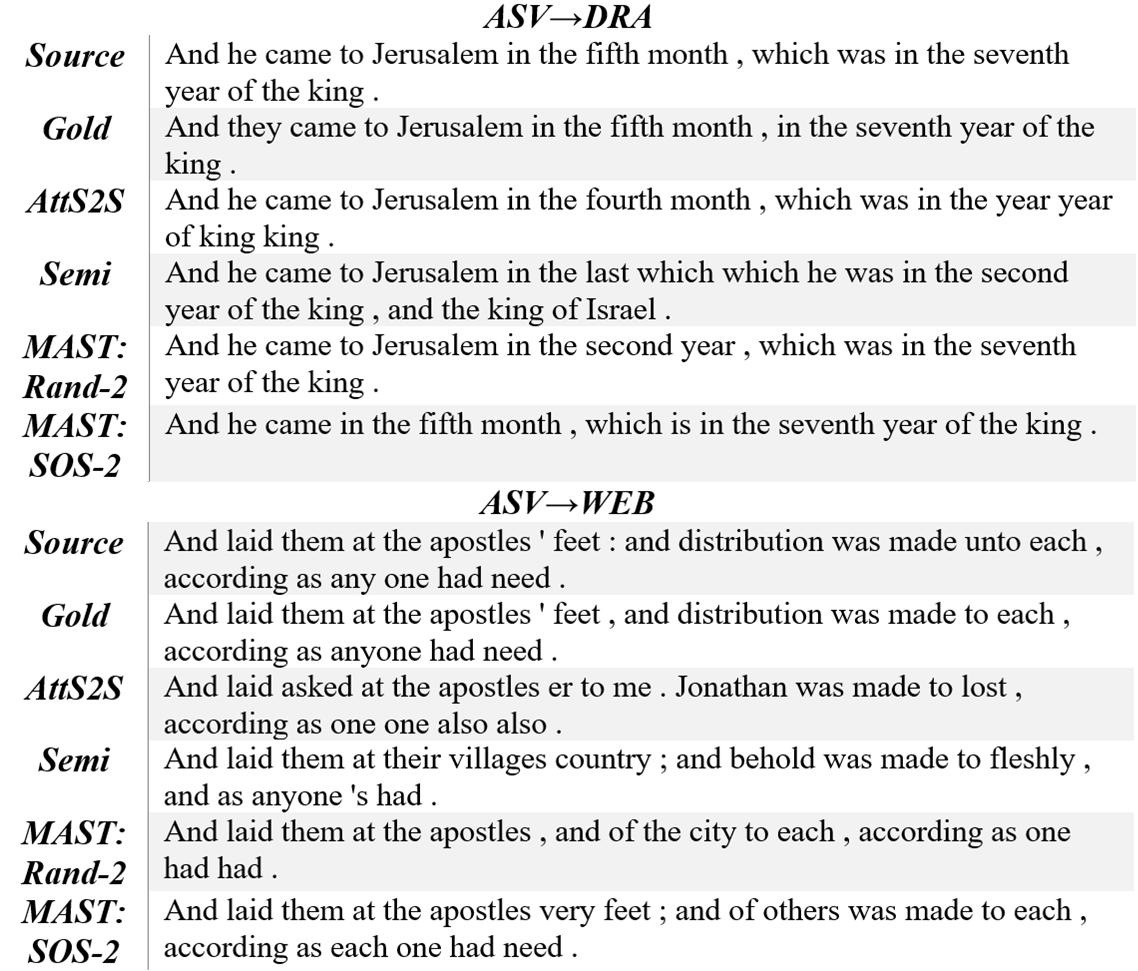}
    \caption{Samples and outputs of different models.}
    \label{figures7}
\end{figure}

\begin{table*}[ht]
\begin{center}

\resizebox{6in}{1.4in}{
\begin{tabular}{|c|c|c|c|c|c|c|c|c|c|c|}
\hline
\multirow{2}* {\bf  Version} & \multirow{2}*{\bf  AttS2S }&\multirow{2}*{\bf   Semi }&\bf   MAST: &\bf   MAST: &\bf   MAST: &\bf   MAST: &\bf   MAST: &\bf   MAST: &\bf   MAST: &\bf   MAST: \\
\bf  ~ &\bf  ~ &\bf  ~ &\bf   Rand-1 &\bf   Rand-2 &\bf   Rand-3 &\bf   Rand-4 &\bf  SOS-1 &\bf   SOS-2 &\bf  SOS-3 &\bf   SOS-4 \\\hline
\bf  BBE & 15.57 & 19.7 & 25.21 & 23.33 & 24.56 & 21.81 & 23.86 & 25.93 & 25.8 & \bf 26.23\\\hline
\bf  DARBY & 29.19 & 32.6 & 31.89 & 38.68 & 38.7  & 31.49 & 40.5 & \bf 40.56 & 40.31 & 39.45\\\hline
\bf  DRA & 15.99 & 21.25 & 17.08 & 20.33 & 18.47 & 28.28 & 27.52 & \bf 30.39 & 29.53 & 29.93\\\hline
\bf  WEB & 40.5 & 41.94 & 43.45 & 40.24 & 35.68 & 35.86 & 43.01 & 45.55 & \bf 46.25 & 44.15\\\hline
\bf  YLT & 15.08 & 19.03 & 20.86 & 20.48 & 18.33 & 17.34 & 22.59 & \bf 24.51 & 23.24 & 23.82\\\hline
\bf  AMP & 10.21 & 12.94 & 10.76 & 15.63 & 13.41 & 15.22 & 14.87 & 15.15 & \bf 15.91 & 15.75\\\hline
\bf  CJB & 12.82 & \bf 14.79 & 11.23 & 10.65 & 12.3 & 12.01 & 12.36 & 12.86 & 12.67 & 14.04\\\hline
\bf  CSB & 8.11 & 12.34 & 12.27 & 18.95 & 15.81 & 19.57 & 19.41 & 19.9 & \bf  20.87 & 20.82\\\hline
\bf  ERV & 2.97 & 4.75 & 7.78 & 8.76 & 8.07 & \bf 8.79 & 7.05 & 8.68 & 8.58 & 8.53\\\hline
\bf  ESV & 18.19 & 24.87 & 26.45 & 25.55 & 28.29 & 29.26 & 32.75 & 33.59 & 34.33 & \bf  34.82\\\hline
\bf  KJ21 & 29.37 & 35.85 & 38.5 & 34.03 & 42.09 & 31.84 & 42.58 & \bf 42.82 & 40.8 & 39.79\\\hline
\bf  MEV & 15.45 & 21.43 & 25.17 & 28.05 & 19.43 & 20.51 & 27.69 & 31.13 & \bf 32.4 & 30.42\\\hline
\bf  NCV & 4.79 & 5.52 & 9.52 & 10.7 & 10.76 & 9.32 & 10.3 & 9.63 & 10.51 & \bf 10.98\\\hline
\bf  NIV & 7.75 & 11.54 & 12.26 & 12.84 & 12.54 & 15.03 & 16.5 & 18.23 & \bf 19.49 & 19.12\\\hline
\bf  NOG & 14.91 & \bf 16.08 & 12.29 & 10.76 & 7.91 & 9.27 & 12.27 & 9.62 & 8.36 & 14.02\\\hline
\bf  Average & 16.06 &19.64 & 20.31 & 21.27 &20.42& 20.37&23.55&24.57&24.60&\bf 24.79
\\\hline

\end{tabular}
}
\end{center}
\caption{\label{tables6}Full BLEU evaluation results for all the version transfer tasks (the last row is the average results).}
\end{table*}

From the comparison of Semi and AttS2S in Table 1 and Table 2, we find that Semi models with DAE and back-translation do help in improving the performance of the basic AttS2S models. And for all the results, our models MAST:Rand-2 and MAST:SOS-2 generally make significant improvements, and MAST:SOS-2 even outperforms AttS2S and Semi by averagely 6.57 BLEU points and 2.94 BLEU points, respectively. MAST:Rand-2 seems to be unsteady, sometimes even a little worse than AttS2S, and much worse than Semi, while MAST:SOS-2 is almost always much better than the baselines models. This shows the effectiveness of our proposed SOS algorithm. But in the results of human evaluation, MAST:SOS-2 is not always significantly better than MAST:Rand-2. Considering the space for choosing neighbors are rather small in these experiments, we extend the dataset with 10 more versions.

\subsection{Extended Experiments}

In this section, we extend the experiments to 15 style (version) transfer tasks, where there are 14 candidate neighbors for each agent. As BLEU has shown good consistency with human evaluation results in previous experiments, we only use BLEU for automatic evaluations. The results are shown in Table 3. As too many numbers make it hard to compare, we also develop color scales in Figure 4 according to the values in Table 3\footnote{The neighbor sets are different from the ones in previous experiments, some results can be different from those in Table 1}.

\begin{figure}[ht!]
    \centering
    \includegraphics[width=3in]{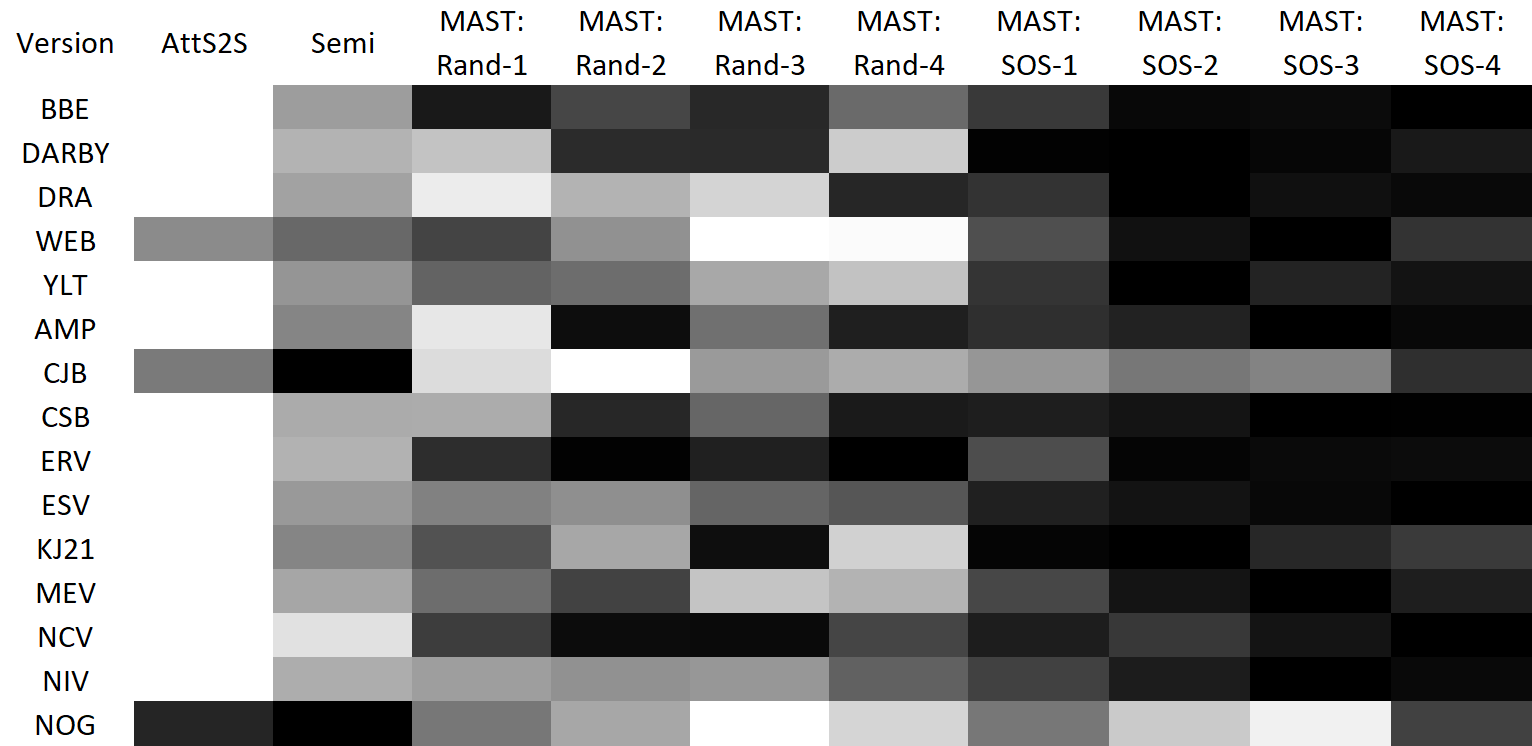}
    \caption{Color scales (from pure white to pure black) of BLEU results. The darker, the better.}
    \label{figures8}
\end{figure}

From Table 3 and Figure 4, we clearly find that the performances of MAST:Rand-k are really random. Although some results of MAST:Rand-k can also achieve the state-of-the-art, many results are even worse than the Semi models they base on. After all, the neighbors are randomly chosen for these agents, so the random results make sense in this way. On the contrary, the results of MAST:SOS-k are much better. From Figure 4, the results of MAST:SOS-k are almost all darker (better) than the others. What's more, we also see that with more neighbors, the results seem to be more steady and significant. For example, MAST:SOS-4 outperforms AttS2S by 8.73 BLEU scores on average, and it also outperforms Semi by 5.15 BLEU scores on average.


\subsection{Further Analysis}

\subsubsection{Influence of Neighbor Candidates}

By comparing the results of MAST:SOS-2 in Table 1 and Table 3, we find that the amount of neighbor candidates has an influence on the performance of our system. We further develop experiments to verify this. In this set of experiments, we vary the versions in our experiments, from the 5 public versions (BBE, DARBY, DRA, WEB, YLT) to all the 15 versions by adding additional versions two by two. As the 5 basic versions exist in all the experiments, we regard the results of MAST:SOS-2 in Table 1 as the initial results, and we calculate the average improvements over initial results for each following experiment of MAST:SOS-2. The trend is shown in Figure 5, where more candidates tend to support our system to get better results.

\subsubsection{Different Data Settings}

We change the number of labeled data pairs from 0.5k to 5k to explore whether the system still provides significant improvements under different data settings. We develop experiments on the 5 basic versions (BBE, DARBY, DRA, WEB, YLT) and measure the average BLEU scores of the major models in this study (AttS2S, Semi and MAST:SOS-2). And we can see in Figure 6, our system performs extremely better than other models in the real low-resource settings, which is very promising for practical uses.

\begin{figure}[t!]
    \centering
    \includegraphics[width=3in]{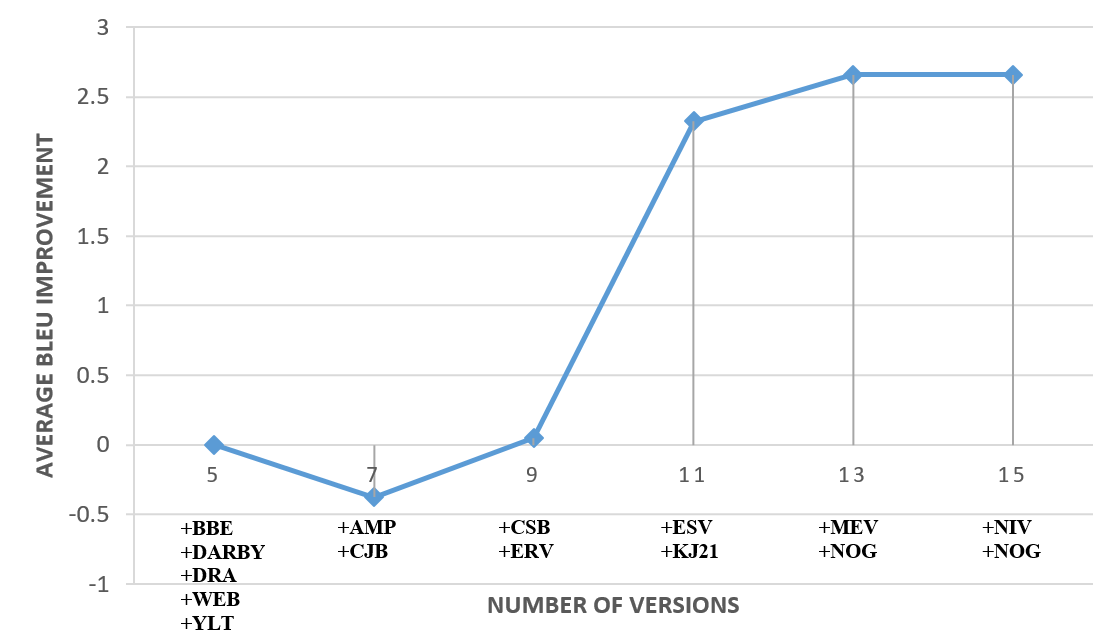} 
    \caption{Changes of performances when adding more candidates.}
    \label{figures9}
\end{figure}

\begin{figure}[t!]
    \centering
    \includegraphics[width=3in]{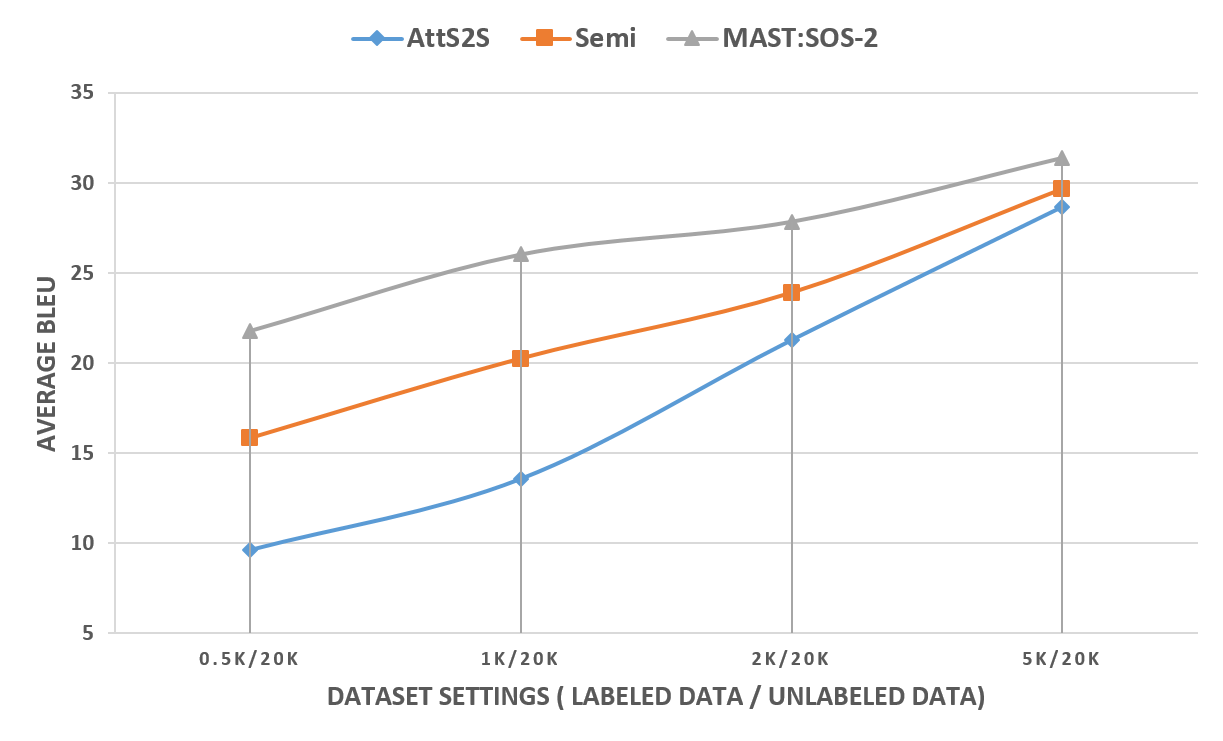}
    \caption{Model performances under different data settings.}
    \label{figures10}
\end{figure}

\section{Conclusion and Future Work}
In this work, we propose a multi-agent system for addressing multiple style transfer tasks with limited labeled data. We design a semi-supervised model to leverage unlabeled data for one-to-one style transfer, and propose MAST for multiple styles transfer. In MAST, SOS picks useful neighbors considering limited resources, and MAT provides controllers for models integration. The design takes account of practicability and expansibility. The comprehensive experiments demonstrate the effectiveness of our system.

In future work, we can explore other algorithms to find neighbor agents more accurately.  Moreover, the controllers we use in MAST have many other options. And basic agents can be improved with other techniques to better support MAST.

\section*{Acknowledgments}
This work was supported by Key Laboratory of Science, Technology and Standard in Press Industry (Key Laboratory of Intelligent Press Media Technology).
Special thanks to Zhiwei Yu for her insightful suggestions. Xiaojun Wan is the corresponding author.

\bibliography{acl2019}

\begin{thebibliography}{27}
\expandafter\ifx\csname natexlab\endcsname\relax\def\natexlab#1{#1}\fi

\bibitem[{Artetxe et~al.(2017)Artetxe, Labaka, Agirre, and
  Cho}]{DBLP:journals/corr/abs-1710-11041}
Mikel Artetxe, Gorka Labaka, Eneko Agirre, and Kyunghyun Cho. 2017.
\newblock \href {http://arxiv.org/abs/1710.11041} {Unsupervised neural machine
  translation}.
\newblock \emph{CoRR}, abs/1710.11041.

\bibitem[{Bahdanau et~al.(2014)Bahdanau, Cho, and Bengio}]{bahdanau2014neural}
Dzmitry Bahdanau, Kyunghyun Cho, and Yoshua Bengio. 2014.
\newblock Neural machine translation by jointly learning to align and
  translate.
\newblock \emph{arXiv preprint arXiv:1409.0473}.

\bibitem[{Bengio et~al.(2013)Bengio, Yao, Alain, and
  Vincent}]{DBLP:conf/nips/BengioYAV13}
Yoshua Bengio, Li~Yao, Guillaume Alain, and Pascal Vincent. 2013.
\newblock Generalized denoising auto-encoders as generative models.
\newblock In \emph{Advances in {NIPS} 2013.}, pages 899--907.

\bibitem[{Carlson et~al.(2017)Carlson, Riddell, and
  Rockmore}]{DBLP:journals/corr/abs-1711-04731}
Keith Carlson, Allen Riddell, and Daniel~N. Rockmore. 2017.
\newblock \href {http://arxiv.org/abs/1711.04731} {Zero-shot style transfer in
  text using recurrent neural networks}.
\newblock \emph{CoRR}, abs/1711.04731.

\bibitem[{Collobert and Weston(2008)}]{DBLP:conf/icml/CollobertW08}
Ronan Collobert and Jason Weston. 2008.
\newblock \href {https://doi.org/10.1145/1390156.1390177} {A unified
  architecture for natural language processing: deep neural networks with
  multitask learning}.
\newblock In \emph{Machine Learning, Proceedings of {(ICML} 2008)}, pages
  160--167.

\bibitem[{Ficler and Goldberg(2017)}]{ficler-goldberg:2017:StyVa}
Jessica Ficler and Yoav Goldberg. 2017.
\newblock Controlling linguistic style aspects in neural language generation.
\newblock In \emph{Proceedings of the Workshop on Stylistic Variation}, pages
  94--104.

\bibitem[{Goodfellow et~al.(2014)Goodfellow, Pouget-Abadie, Mirza, Xu,
  Warde-Farley, Ozair, Courville, and Bengio}]{goodfellow2014generative}
Ian Goodfellow, Jean Pouget-Abadie, Mehdi Mirza, Bing Xu, David Warde-Farley,
  Sherjil Ozair, Aaron Courville, and Yoshua Bengio. 2014.
\newblock Generative adversarial nets.
\newblock In \emph{Advances in neural information processing systems}, pages
  2672--2680.

\bibitem[{Gorodetskii(2012)}]{Gorodetskii2012}
V.~I. Gorodetskii. 2012.
\newblock \href {https://doi.org/10.1134/S1064230712020062} {Self-organization
  and multiagent systems: Ii. applications and the development technology}.
\newblock \emph{Journal of Computer and Systems Sciences International},
  51(3):391--409.

\bibitem[{Hochreiter and Schmidhuber(1997)}]{hochreiter1997long}
Sepp Hochreiter and J{\"u}rgen Schmidhuber. 1997.
\newblock Long short-term memory.
\newblock \emph{Neural computation}, 9(8):1735--1780.

\bibitem[{Jhamtani et~al.(2017)Jhamtani, Gangal, Hovy, and
  Nyberg}]{jhamtani-EtAl:2017:StyVa}
Harsh Jhamtani, Varun Gangal, Eduard Hovy, and Eric Nyberg. 2017.
\newblock Shakespearizing modern language using copy-enriched sequence to
  sequence models.
\newblock In \emph{Proceedings of the Workshop on Stylistic Variation}, pages
  10--19.

\bibitem[{Kiddon et~al.(2015)Kiddon, Ponnuraj, Zettlemoyer, and
  Choi}]{DBLP:conf/emnlp/KiddonPZC15}
Chlo{\'{e}} Kiddon, Ganesa~Thandavam Ponnuraj, Luke Zettlemoyer, and Yejin
  Choi. 2015.
\newblock Mise en place: Unsupervised interpretation of instructional recipes.
\newblock In \emph{Proceedings of {EMNLP} 2015}, pages 982--992.

\bibitem[{Lample et~al.(2017)Lample, Denoyer, and
  Ranzato}]{DBLP:journals/corr/abs-1711-00043}
Guillaume Lample, Ludovic Denoyer, and Marc'Aurelio Ranzato. 2017.
\newblock Unsupervised machine translation using monolingual corpora only.
\newblock \emph{CoRR}, abs/1711.00043.

\bibitem[{Lin and Walker(2017)}]{lin-walker:2017:StyVa}
Grace Lin and Marilyn Walker. 2017.
\newblock Stylistic variation in television dialogue for natural language
  generation.
\newblock In \emph{Proceedings of the Workshop on Stylistic Variation}, pages
  85--93.

\bibitem[{Luong et~al.(2015)Luong, Pham, and
  Manning}]{DBLP:conf/emnlp/LuongPM15}
Thang Luong, Hieu Pham, and Christopher~D. Manning. 2015.
\newblock Effective approaches to attention-based neural machine translation.
\newblock In \emph{Proceedings of {EMNLP} 2015}, pages 1412--1421.

\bibitem[{Nisioi et~al.(2017)Nisioi, Stajner, Ponzetto, and
  Dinu}]{DBLP:conf/acl/NisioiSPD17}
Sergiu Nisioi, Sanja Stajner, Simone~Paolo Ponzetto, and Liviu~P. Dinu. 2017.
\newblock \href {https://doi.org/10.18653/v1/P17-2014} {Exploring neural text
  simplification models}.
\newblock In \emph{Proceedings of {ACL} 2017}, pages 85--91.

\bibitem[{Papineni et~al.(2002)Papineni, Roukos, Ward, and
  Zhu}]{papineni2002bleu}
Kishore Papineni, Salim Roukos, Todd Ward, and Wei-Jing Zhu. 2002.
\newblock Bleu: a method for automatic evaluation of machine translation.
\newblock In \emph{Proceedings of {ACL} 2002}, pages 311--318.

\bibitem[{Prabhumoye et~al.(2018)Prabhumoye, Tsvetkov, Salakhutdinov, and
  Black}]{DBLP:conf/acl/TsvetkovBSP18}
Shrimai Prabhumoye, Yulia Tsvetkov, Ruslan Salakhutdinov, and Alan~W. Black.
  2018.
\newblock \href {https://aclanthology.info/papers/P18-1080/p18-1080} {Style
  transfer through back-translation}.
\newblock In \emph{Proceedings of {ACL} 2018}, pages 866--876.

\bibitem[{Schuster and Paliwal(1997)}]{schuster1997bidirectional}
Mike Schuster and Kuldip~K Paliwal. 1997.
\newblock Bidirectional recurrent neural networks.
\newblock \emph{IEEE Transactions on Signal Processing}, 45(11):2673--2681.

\bibitem[{Sennrich et~al.(2016)Sennrich, Haddow, and
  Birch}]{DBLP:conf/acl/SennrichHB16}
Rico Sennrich, Barry Haddow, and Alexandra Birch. 2016.
\newblock Improving neural machine translation models with monolingual data.
\newblock In \emph{Proceedings of {ACL} 2016}.

\bibitem[{Shen et~al.(2017)Shen, Lei, Barzilay, and
  Jaakkola}]{DBLP:conf/nips/ShenLBJ17}
Tianxiao Shen, Tao Lei, Regina Barzilay, and Tommi~S. Jaakkola. 2017.
\newblock Style transfer from non-parallel text by cross-alignment.
\newblock In \emph{In Processing of {NIPS} 2017}, pages 6833--6844.

\bibitem[{Sutskever et~al.(2014)Sutskever, Vinyals, and
  Le}]{sutskever2014sequence}
Ilya Sutskever, Oriol Vinyals, and Quoc~V Le. 2014.
\newblock Sequence to sequence learning with neural networks.
\newblock In \emph{Advances in neural information processing systems}, pages
  3104--3112.

\bibitem[{Xu et~al.(2018)Xu, Sun, Zeng, Ren, Zhang, Wang, and
  Li}]{DBLP:journals/corr/abs-1805-05181}
Jingjing Xu, Xu~Sun, Qi~Zeng, Xuancheng Ren, Xiaodong Zhang, Houfeng Wang, and
  Wenjie Li. 2018.
\newblock Unpaired sentiment-to-sentiment translation: {A} cycled reinforcement
  learning approach.
\newblock In \emph{Proceedings of the ACL2018.}

\bibitem[{Xu(2017)}]{xu:2017:StyVa}
Wei Xu. 2017.
\newblock From shakespeare to twitter: What are language styles all about?
\newblock In \emph{Proceedings of the Workshop on Stylistic Variation}, pages
  1--9.

\bibitem[{Xu et~al.(2012)Xu, Ritter, Dolan, Grishman, and
  Cherry}]{xu2012paraphrasing}
Wei Xu, Alan Ritter, Bill Dolan, Ralph Grishman, and Colin Cherry. 2012.
\newblock Paraphrasing for style.
\newblock In \emph{COLING}, pages 2899--2914.

\bibitem[{Yang et~al.(2018{\natexlab{a}})Yang, Chen, Wang, and
  Xu}]{zhenyang2018}
Zhen Yang, Wei Chen, Feng Wang, and Bo~Xu. 2018{\natexlab{a}}.
\newblock Unsupervised neural machine translation with weight sharing.
\newblock In \emph{Proceedings of the ACL2018.}

\bibitem[{Yang et~al.(2018{\natexlab{b}})Yang, Hu, Dyer, Xing, and
  Berg{-}Kirkpatrick}]{DBLP:conf/nips/YangHDXB18}
Zichao Yang, Zhiting Hu, Chris Dyer, Eric~P. Xing, and Taylor
  Berg{-}Kirkpatrick. 2018{\natexlab{b}}.
\newblock Unsupervised text style transfer using language models as
  discriminators.
\newblock In \emph{Advances in {NIPS} 2018}.

\bibitem[{Yu et~al.(2017)Yu, Zhang, Wang, and Yu}]{DBLP:conf/aaai/YuZWY17}
Lantao Yu, Weinan Zhang, Jun Wang, and Yong Yu. 2017.
\newblock Seqgan: Sequence generative adversarial nets with policy gradient.
\newblock In \emph{Proceedings of {AAAI} 2017}, pages 2852--2858.

\end{thebibliography}
 \bibliographystyle{acl_natbib} 
\end{document}